\documentclass[sigconf]{acmart}

\AtBeginDocument{%
  }

\setcopyright{acmcopyright}
\copyrightyear{2025}
\acmYear{2025}
\acmDOI{XXXXXXX.XXXXXXX}

\acmConference[AICCC]{The 2025 Annual International Conference on Cognitive Computing (AICCC)}{Dec 20--22,
  2025}{Tokyo, JP}

\acmISBN{979-8-4007-1889-2}

\begin{document}


\title{SPLite Hand: Sparsity-Aware Lightweight 3D Hand Pose Estimation}

\author{Yeh Keng Hao}
\authornote{First author and primary contributor.}
\affiliation{
  \institution{National Tsing Hua University}
  \country{Taiwan}
}
\email{haoyeh@gapp.nthu.edu.tw}

\author{Hsu Tzu Wei}
\authornote{Second author.}
\affiliation{
  \institution{National Tsing Hua University}
  \country{Taiwan}
}
\email{welly9166@gmail.com}

\author{Sun Min}
\authornote{Third author. Project advisor.}
\affiliation{
  \institution{National Tsing Hua University}
  \country{Taiwan}
}
\email{sunmin@ee.nthu.edu.tw}


\begin{abstract}

With the increasing ubiquity of AR/VR devices, the deployment of deep learning models on edge devices has become a critical challenge. These devices require real-time inference, low power consumption, and minimal latency. Many framework designers face the conundrum of balancing efficiency and performance.We design a light framework that adopts an encoder-decoder architecture and introduces several key contributions aimed at improving both efficiency and accuracy. We apply sparse convolution on a ResNet-18 backbone to exploit the inherent sparsity in hand pose images, achieving a 42\% end to end efficiency improvement.Moreover, we propose our SPLite decoder. This new architecture significantly boosts the decoding process's frame rate by 3.1$\times$ on the Raspberry Pi 5, while maintaining accuracy on par. To further optimize performance, we apply quantization-aware training, reducing memory usage while preserving accuracy (PA-MPJPE increases only marginally from 9.0\,mm to 9.1\,mm on FreiHAND). Overall, our system achieves a 2.98$\times$ speed-up on a Raspberry Pi 5 CPU (BCM2712 quad-core Arm A76 processor). Our method is also evaluated on compound benchmark datasets, demonstrating comparable accuracy to state-of-the-art approaches while significantly enhancing computational efficiency.
\end{abstract}




\maketitle

\section{Introduction}

3D hand pose estimation from a single 2D input is vital for applications in robotics, VR, and AR. The primary challenge is balancing high accuracy with computational efficiency, especially for devices with limited resources.

Rather than designing transformer-based architectures~\cite{lin2021endtoendhumanposemesh, liu2024keypoint,guo2018stacked} to achieve higher accuracy, recent research has focused on building cost-effective, lightweight neural networks with reduced computational complexity so as to deploy on edge devices. Works such as MobileNet~\cite{howard2017mobilenetsefficientconvolutionalneural,sandler2019mobilenetv2invertedresidualslinear}, ShuffleNet~\cite{zhang2017shufflenetextremelyefficientconvolutional}, and GhostNet~\cite{han2020ghostnet} introduced efficient operators such as depthwise and group convolution. However, these methods often suffer from increased memory access costs and suboptimal memory utility.

A common pipeline for single-view hand pose estimation typically consists of three stages: 2D encoding, 2D-to-3D feature lifting, and 3D decoding. Most of the computation occurs during the 2D encoding phase. Lin \textit{et al.}~\cite{lim2020mobilehand} directly use MobileNet to achieve a higher frame rate~\cite{howard2017mobilenetsefficientconvolutionalneural,sandler2019mobilenetv2invertedresidualslinear}, which struggles with predicting occlusion and complex scenarios. It also can't apply custom operations like sparse convolution, while others\cite{choi2021staticfeaturestemporallyconsistent,li2021human} introduce customized blocks to extract 2D hand keypoints \( \mathbf{J}_{2D} \in \mathbb{R}^{21 \times 2} \). Over time, several techniques have been proposed to make deep networks more compact, such as pruning~\cite{he2023structured,li2020cnnpruner}, low-bit quantization~\cite{wang2018two,choukroun2019low}, and knowledge distillation~\cite{fukuda2017efficient,aghli2021combining}. However, the effectiveness of these methods is often upper-bounded by the quality of the pre-trained models used as baselines.

To reduce redundancy in feature maps and improve efficiency, Chen \textit{et al.}~\cite{chen2023rundontwalkchasing} proposed a heuristic partial-channel design, FasterNet, which operates only on selected channels. Inspired by this, we adopt a 3-stage architecture—2D encoder, 2D-3D feature lifting, and 3D decoder, termed SPLite, to predict both 3D keypoints and mesh as illustrated in Figure~\ref{fig:framework}.

In the 2D encoding stage, We use a proprietary algorithm to generate the edge image, as shown in Figure~\ref{fig:wilddata2}, and send it to the early-fusion block, which fuse two types of edge modalities. This avoids the need for modifying the backbone while naturally introducing sparsity. Sparse convolution, unlike dense convolution, operates only on non-zero elements, reducing both memory usage and computational load. Inspired by Zhang \textit{et al.}~\cite{zhang2025high}, We leverage sparse convolution to focus on high-intensity edge regions, improving inference speed by 42\% without sacrificing accuracy. This is particularly beneficial for deploying the model on edge devices with limited power consumption.

For the 2D-to-3D lifting stage, we adopt a simplified pose-to-vertex module based on Chen \textit{et al.}'s lifting module~\cite{chen2022mobreconmobilefriendlyhandmesh}, which uses a lifting matrix to project the 2D sequential feature maps into 3D representations.

In 3D decoders, \textit{Graph Convolutional Networks} (GCNs)~\cite{chen2021channel,kipf2017semisupervisedclassificationgraphconvolutional} are a specialized type of neural network that can effectively analyze irregular data structure, such as meshes. GCNs are often used to handle these data types by learning meaningful representations for each node. GCNs achieve this by aggregating and combining features from a node's neighbors, which allows them to capture both the node's individual properties and its structural context within the graph. 

Although many researchers have proposed improvements to this operation---for example, Gong \textit{et al.} proposed SpiralNet++~\cite{gong2019spiralnetfasthighlyefficient} to enhance SpiralNet~\cite{bouritsas2019neural}---these new architectures often slow down the forward pass because the vertex sampling process is computationally expensive.

Large pre-trained networks are effective for vision tasks but face memory and computation limitations on embedded systems. While model quantization to low-bit precision can compress these networks, it often sacrifices accuracy. Wang \textit{et al.}~\cite{ashkboos2024efqatefficientframeworkquantizationaware,wang2018two} have shown that Quantization-aware training (QAT) can preserve accuracy. We first train a floating-point baseline, then apply QAT, and convert the model to ONNX for deployment. An adjustable learning rate compensates for reduced accuracy. On benchmark dataset\cite{zimmermann2019freihanddatasetmarkerlesscapture}, our quantized model reduces size by 75\% (from 72MB to 18MB) with only a 0.1mm drop in PA-MPJPE (from 9.0mm to 9.1mm).

Finally, we introduce a new multimodal hand-object interaction dataset with dual-view perspectives. It includes RGB, grayscale, and edge modalities across 100 object categories and manipulation actions.

\subsection*{Our key contributions}
\begin{itemize}
  \item We introduce a sparse data transformation pipeline and apply sparse convolution to our encoder, achieving a 42\% speed-up.
  \item We propose the \textbf{SPLite decoder}, a hardware-friendly graph-based decoder. It accelerates inference by up to $3.1\times$ (vs. \textit{MobRecon-tailored}~\cite{chen2022mobreconmobilefriendlyhandmesh}) and 65\% (vs. \textit{ResNet18}~\cite{he2016deep}) on Raspberry Pi 5.

  \item We apply quantization-aware training to compress our model from 72MB to 18MB ($\downarrow$75\% ), with only a 0.1mm accuracy drop in PA-MPJPE.
  
  \item We present a diverse and challenging hand-object interaction dataset, featuring multi-modal data across 100 object categories and manipulation actions for audience to evaluate their models.
\end{itemize}

\section{Related Work}

\begin{figure*}[t]
    \centering
    \includegraphics[width=0.8\textwidth]{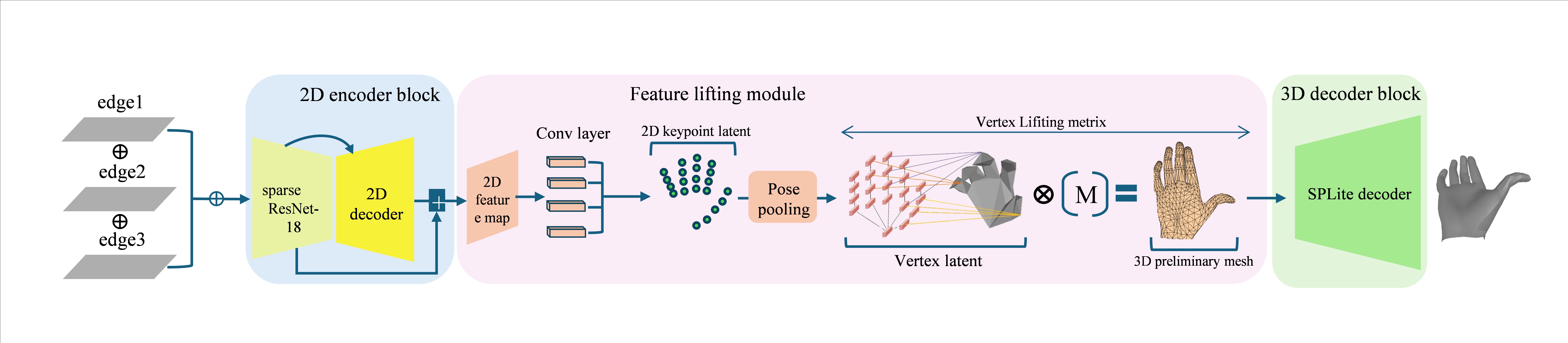}
    \caption{our efficient pipeline begins with an RGB image being converted to grayscale, from which a proprietary algorithm extracts an edge image. A customized backbone called Sparse ResNet-18 then uses sparse convolutional blocks to generate 2D feature maps and predicted depth values. To improve the robustness of the depth prediction, a second data stream is introduced during training to account for perspective ambiguity. Finally, a calibrated camera intrinsic transformation module converts the 2D feature maps from camera coordinates to real-world coordinates for the lightweight SPLite decoder, which ultimately predicts a 3D hand mesh.}
    \label{fig:framework}
\end{figure*}

\subsection*{Lightweight Networks}
Lightweight architectures have been extensively studied to reduce computational burden, particularly for deployment on resource-constrained platforms. MobileNet~\cite{howard2017mobilenetsefficientconvolutionalneural,sandler2019mobilenetv2invertedresidualslinear} utilizes depthwise separable convolutions (depth-wise and pointwise convolutions) to reduce computation. ShuffleNet~\cite{zhang2017shufflenetextremelyefficientconvolutional} introduces a channel shuffle mechanism to enhance information flow and reduce latency. GhostNet~\cite{han2020ghostnet} further leverages depthwise and group convolutions to extract spatial features more efficiently.
Chen \textit{et al.} ~\cite{chen2022mobreconmobilefriendlyhandmesh} proposed a lightweight model design as our baseline, After replacing the encoding and decoding modules of our baseline model with a customized version, we achieved a significant breakthrough across several performance metrics.
While these methods have demonstrated effectiveness in reducing computation, many of them incur increased memory access costs and inefficient memory utility—limitations that become critical on edge devices. Moreover, many "lightweight" models are only validated on high-end GPUs, rather than actual edge platforms. In contrast, our SPLite operator shows efficiency and memory improvements on real-world edge hardware such as the Raspberry Pi 5, and even outperforming baselines in several use cases, as shown in Figure~\ref{fig:wilddata1}.

\subsection*{Sparse Convolution}
Convolutional neural networks (CNNs) have achieved significant success across various visual tasks, including 2D/3D hand pose estimation. However, the dense nature of traditional convolution operations leads to high computational overhead, especially for high-dimensional outputs like 3D meshes.
To address this, libraries such as Minkowski Engine~\cite{minkowski2019} and TorchSparse~\cite{tang2023torchsparse++} have been developed to enable efficient sparse convolution for 3D point cloud processing. In 2D domains, Hsieh \textit{et al.}~\cite{lin2023sparse} introduced sparse convolution for body pose estimation by leveraging motion vector cues. However, this approach is ineffective when the subject is still, as motion vectors provide no useful information. To overcome this limitation, we adopt a different fusion strategy and incorporate an accelerating module ,termed SPLite, within our decoding process. Inspired by these works, we propose to process input images into edge-enhanced sparse representations and then apply sparse convolution. Our experiments show that when the input sparsity reaches a mean of 86\%, our sparse convolution design improves inference speed by 42\%.

\subsection*{Hand-Object Interaction Datasets}
Hand-object Interaction (HOI) datasets play a central role in training models for realistic manipulation tasks. Several well-known datasets have been introduced, such as HO-3D~\cite{hampali2021ho}, HOI4D~\cite{liu2024hoi4d4degocentricdataset}, and HOT3D~\cite{banerjee2025hot3dhandobjecttracking}, providing benchmarks for joint hand-object tracking and pose estimation. However, many of these datasets are limited in either object diversity, interaction complexity, or sensory modality.

Our dataset addresses these limitations by incorporating synchronized RGB, grayscale, and edge modalities, with over 100 object categories and manipulation types. It provides dual-view perspectives, enabling richer supervision and facilitating better generalization across interaction scenarios.

%% identificati

\section{Our Method}

We aim to accelerate our model without sacrificing accuracy. To achieve this, we design an efficient pipeline consisting of several stages: early fusion, 2D encoding, feature lifting, and 3D decoding. Figure~\ref{fig:framework} illustrates the holistic architecture. Our objective is to predict a 3D hand mesh from a monocular RGB image.

First, the input RGB image is converted to grayscale, and an edge map is produced using our proprietary edge-detection algorithm—methods such as the Canny\cite{canny1986computational} or Sobel detectors\cite{sobel1968isotropic} yield comparably effective results. We employ a customized backbone, \textit{Sparse ResNet-18}, which leverages the inherent sparsity of the input data by applying sparse convolutional blocks.

In the 2D encoding stage, our network produces preliminary 2D feature maps in camera coordinates $(u_i, v_i),~i=1\ldots N$, along with predicted depth values. Since monocular depth estimation is ill-posed due to perspective ambiguity, we enhance the robustness of depth prediction during training by introducing a second-view data stream.

The 2D feature maps are then transformed from camera coordinates into real-world coordinates using a calibrated camera intrinsic transformation module. This transformation ensures spatial consistency before feeding the features into our lightweight SPLite decoder.

\subsection{Early Fusion}

As shown in Figure~\ref{fig:framework}, we first process the RGB input images into edge maps. The pixel values of these edge images range from 0 to 255, where higher values indicate stronger edge intensity. A fusion module then combines two types of edge representations into a sandwich-like structure, which is passed to the 2D encoding module.

\subsection{2D Encoding}

We adopt ResNet-18 as the backbone for our 2D encoder due to its lightweight architecture and widespread deployment on edge devices. The sparsity of our input data allows us to integrate Minkowski sparse convolutions~\cite{minkowski2019}, resulting in a 36\%--42\% speed-up on Intel\textregistered~Core\texttrademark~i9-9980XE CPU. Also, We leverage \(7 \times 7\) kernel, a larger convolutional kernel size, to capture broader spatial hand pose context and enhance feature representation.
On average, our dataset exhibits an input sparsity of 89.2\%, enabling sparse convolution to deliver performance comparable to dense convolution while drastically reducing the inference time. This strategy has also proven successful in domains such as 3D point cloud processing and 2D character recognition.

\subsection{2D-to-3D Feature Lifting}
To lift 2D features into a 3D representation, we utilize a feature elevation component that connects the 2D and 3D realms via a sequence of optimized procedures. This component commences with heatmap-guided estimation to forecast 2D landmark locations from the feature grid, proceeds to gather synchronized features. We associate it with a predicted depth value \(d\), which allows us to compute the corresponding 3D point \((x, y, z)\) in real-world coordinates. The process consists of the following steps:

\begin{itemize}

\item \textbf{Heatmap-regression}: \\
The 2D feature grid generated by the encoder is handled to estimate 2D landmark locations. This entails producing probability distributions over a mid-resolution feature grid 
\[
L_{l} = \{L_{l}^{j}\}_{j=1}^{M},
\]
where $M$ is the keypoint count, 21 keypoints in a hand.

\item \textbf{Pose Pooling}: \\
Leveraging the forecasted 2D landmark locations, we derive alignment-synchronized features from the encoded 2D feature grid $F_{g}$:
\[
F_{p} = \big[\,F_{g}(L_{l}^{j})\,\big]_{j=1}^{M}.
\]
This produces a collection of $M$ feature vectors $F_{a}\in\mathbb{R}^{M\times C_{g}}$, where $C_{g}$ is the feature dimension, all spatially matched to the 2D landmarks.

\item \textbf{Pose-to-Vertex Lifting}: \\
To convert the 2D-synchronized features into 3D mesh point features, we apply a trainable lifting matrix on a subsampled reference mesh. The mapping is executed as:
\[
F_{\text{mesh}} = M \cdot F_{p}
\]
where $M$ denotes the trained Matrix that translates landmark features to subsampled mesh features, upholding conceptual and structural integrity without depending on overt spatial transformations.

\end{itemize}

This transformation embeds geometric consistency into the feature map, which is crucial for accurate 3D hand mesh reconstruction. The resulting 3D-aware features, now spatially aligned in real-world coordinates, are then passed to our SPLite decoder for further processing.

This transformation embeds geometric consistency into the feature map, which is crucial for accurate 3D hand mesh reconstruction. The resulting 3D-aware features, now spatially aligned in real-world coordinates, are then passed to our SPLite decoder for further processing.

\subsection{3D Decoding with SPLite Decoder}

We propose a novel SPLite graph operator inspired by Spiral Convolution~\cite{gong2019spiralnetfasthighlyefficient}, tailored for efficient hardware deployment. Traditional spiral-based graph convolutions suffer from high latency due to sequential vertex traversal. To overcome this, we introduce a parallelized sampling strategy that restructures the vertex indexing process to support \textit{SIMD} (Single Instruction, Multiple Data) operations at the compiler level. This enables simultaneous processing of multiple vertex indices, accelerating vertex sampling process and improving CPU utilization.

Specifically, the SPLite module parallelly indexes the vertex latent, using only a quarter of the channels for the decoding convolution. This technique efficiently reduces redundant computation and memory access time while preserving model accuracy. These two features—parallel indexing and partial channel convolution are the foundation of our SPLite decoder.

Integrated together, these optimizations yield significant performance improvements. Specifically, our SPLite decoder achieves a 28\% speed-up over the MobRecon-tailored~\cite{chen2022mobreconmobilefriendlyhandmesh} baseline, and a 65\% speed-up compared to MobRecon~\cite{chen2022mobreconmobilefriendlyhandmesh} with ResNet-18~\cite{he2016deep} on the Raspberry Pi 5 CPU. Ablation studies indicate that the partial channel decoding alone contributes up to a $2.84\times$ speed increase over full-channel decoding.
\section{Loss Functions}

To train the model and ensure accurate 3D reconstruction, we employ various loss functions that optimize different aspects of the 3D pose and shape estimation process. These loss functions typically focus on the following objectives:

\subsection{Reprojection Loss}

We adopt reprojection loss to minimize the difference between the 2D projections of the estimated 3D keypoints and the actual 2D keypoints observed in the image. It is formulated as:

\[
L_{\text{reproj}} = \sum_{i} \| \mathbf{P}_{i}^{\text{2D}} - \hat{\mathbf{P}}_{i}^{\text{2D}} \|_2^2
\]

Where:
\begin{itemize}
    \item \(\mathbf{P}_{i}^{\text{2D}}\) is the ground truth 2D position of the \(i\)-th keypoint.
    \item \(\hat{\mathbf{P}}_{i}^{\text{2D}}\) is the predicted 2D position of the \(i\)-th keypoint, which is obtained by projecting the predicted 3D points into the 2D image space using the intrinsic matrix \(\mathbf{K}\).
\end{itemize}

The goal of this loss is to reduce the Euclidean distance between the predicted and ground truth 2D points, ensuring that the projected 3D points match the 2D observations as closely as possible.

\subsection{3D Pose Loss}

After our model predict 3D keypoint position. 3D pose losses focus on the difference between the predicted 3D keypoints and the ground truth 3D keypoints. It is  computed with the L2 norm:

\[
L_{\text{pose}} = \sum_{i} \| \hat{\mathbf{P}}_{i}^{\text{3D}} - \mathbf{P}_{i}^{\text{3D}} \|_2^2
\]

Where:
\begin{itemize}
    \item \(\mathbf{P}_{i}^{\text{3D}}\) is the ground truth 3D position of the \(i\)-th keypoint.
    \item \(\hat{\mathbf{P}}_{i}^{\text{3D}}\) is the predicted 3D position of the \(i\)-th keypoint.
\end{itemize}

The goal of 3D keypoint loss function is to minimize the error between the predicted 3D points and their ground truth counterparts.

\subsection{Depth Loss}
Estimating depth from a single RGB image is inherently error-prone. We obtain ground-truth depth data from the Intel® RealSense™ D435 camera. The depth loss is defined as:

\[
L_{\text{depth}} = \sum_{i} \| d_i - \hat{d}_i \|_2^2
\]

Where:
\begin{itemize}
    \item \(d_i\) is the ground truth depth of the \(i\)-th keypoint.
    \item \(\hat{d}_i\) is the predicted depth of the \(i\)-th keypoint.
\end{itemize}

This loss directly affects the reconstruction of the 3D structure.

\subsection{Smoothness Loss}

To improve the smoothness of the reconstructed 3D mesh, a smoothness loss is applied to the 3D vertices, ensuring that neighboring vertices in the mesh are not overly distorted. It is formulated as:

\[
L_{\text{smooth}} = \sum_{i} \| \hat{\mathbf{V}}_i - \hat{\mathbf{V}}_j \|_2^2
\]

Where:
\begin{itemize}
    \item \(\hat{\mathbf{V}}_i\) and \(\hat{\mathbf{V}}_j\) are neighboring vertices in the 3D mesh.
\end{itemize}

Smooth loss encourages spatial consistency between neighboring points, which helps in generating more realistic and smooth 3D meshes.

\subsection{Aggregation Loss}

The aggregation loss used for training is a weighted sum of the individual losses:

\[
L_{\text{aggregation}} = \lambda_1 L_{\text{reproj}} + \lambda_2 L_{\text{pose}} + \lambda_3 L_{\text{depth}} + \lambda_4 L_{\text{smooth}}
\]

Where \(\lambda_1, \lambda_2, \lambda_3, \lambda_4\) are hyperparameters that control the relative importance of each loss term. 

\section{Experiments}
\subsection{Experiment details}

We use an input resolution of \(128 \times 128\) for all experiments. The model is trained on images resized to this resolution to standardize input dimensions and maintain computational efficiency. For data augmentation, we apply random cropping, flipping, and color jittering during training to improve model generalization. The input data consists of both real and synthetic hand pose datasets, with the latter being uniformly distributed over 1520 poses and 216 viewpoints, ensuring comprehensive coverage of diverse hand configurations and viewpoints.
The Adam optimizer is used for training with an initial learning rate of \(10^{-3}\), which is reduced 10 $\times$ after the 30th epoch. We use a mini-batch size of 32, and the models are trained for a total of 38 epochs. The hyperparameters for our model follow the configuration specified in the baseline work \cite{chen2022mobreconmobilefriendlyhandmesh}, ensuring consistency and fairness in comparison.

Table~\ref{tab:SparseConv} presents the results of our unit test. With 80\% sparsity, the model achieves 39 fps on ResNet-18, delivering a speed-up of at least 2.1x over the dense convolution, whose frame rate remains constant. Our results demonstrate that sparse convolution effectively leverages sparse data derived from RGB images to achieve a 2$\times$ to 3$\times$ speed-up in frame rate. In contrast, dense convolution processes both sparse and dense inputs uniformly, showing no improvement in performance.
 \begin{table}[h!]
\centering
\caption{Frame rate of sparse/dense convolution. FPS increases with higher input sparsity in sparse convolution.}
\label{tab:SparseConv}
\begin{tabular}{l|cccc}
    \toprule
    Architecture        & 80\%  & 85\%      &90\%      \\
    \midrule
    ResNet-18        & 18   & 18       & 19     \\
    Sparse ResNet-18 & \textbf{39}   & \textbf{42}       & \textbf{51}    \\
    \midrule
    ResNet-50        & 4    & 3       & 4      \\
    Sparse ResNet-50 & \textbf{16}   & \textbf{18}     & \textbf{20}     \\
    
    \bottomrule
\end{tabular}
\end{table}

Table~\ref{tab:QAT} illustrates our model's ability to generalize on multimodal data. The results of applying Quantization-Aware Training (QAT) are shown in Table~\ref{tab:QAT}. Accuracy, measured on RGB images, remains comparable, with PA-MPJPE largely unchanged before and after QAT. Additionally, we train our model with multimodal data by transforming the FreiHAND dataset \cite{zimmermann2019freihanddatasetmarkerlesscapture} into edge images and using our fusion module. Testing on edge images with and without QAT yields PA-MPJPE values of 12.3mm and 11.2mm, respectively, demonstrating the model’s robustness across different modalities.

\subsection{Comparison with existing method}
We utilize compound benchmark dataset. Rendered Hand Pose Dataset (RHD) \cite{zimmermann2017learningestimate3dhand} consists of
41,258 and 2,728 synthetic hand data for training and
testing on hand pose estimation, respectively. And FreiHAND~\cite{zimmermann2019freihanddatasetmarkerlesscapture} contains 130,240 training images and
3,960 evaluation samples.\\

\begin{table}[h!]
\centering
\caption{Comparison of lightweight model speed, parameter and accuracy on the benchmark dataset \cite{zimmermann2019freihanddatasetmarkerlesscapture} First and Second best results are represented as \textcolor{red}{RED} and \textcolor{blue}{BLUE} .}
\label{tab:E2E_rasberry_pi}
\begin{tabular}{l|ccc}
    \toprule
    Method & FPS$\uparrow$ & Params (M)$\downarrow$ & PJ (mm)$\downarrow$ \\
    \midrule
    METRO~\cite{lin2021end}                              & 2            & 102               & \textcolor{blue}{6.7} \\ 
    MobileHand~\cite{lim2020mobilehand}                  & \textcolor{red}{35} & \textcolor{blue}{3.2}                &13  \\ 
    SimpleHand~\cite{zhou2024simplebaselineefficienthand} & 4           & \textcolor{red}{1.9}      &\textcolor{red}{5.8}  \\
    FastMETRO~\cite{cho2022cross}                          & 3          & 25                &6.5  \\ 
    MeshGraphomer~\cite{lin2021meshgraphormer}                          & 1        &98                 &6.3  \\ 
    MobRecon(baseline)~\cite{chen2022mobreconmobilefriendlyhandmesh} & 6  & 8.16   & 9.2    \\
    
    \midrule 
    Ours                                                    & \textcolor{blue}{15}     & 8.79           & 9.1    \\
    \bottomrule
\end{tabular}
\end{table}

%%%  here
MobRecon \cite{chen2022mobreconmobilefriendlyhandmesh} serves as our baseline because it represents one of the most efficient and lightweight architectures proposed in recent years at a top-tier conference, offering a strong balance between accuracy and computational cost. Its design makes it especially suitable for mobile and resource-constrained environments, which aligns with the goals of our work.
Table~\ref{tab:E2E_rasberry_pi} presents a performance comparison between our proposed model and several state-of-the-art lightweight models. To highlight the practical efficiency of our model, we conducted a real-world evaluation on a Raspberry Pi 5 CPU.

Our model achieves a 3x improvement in inference speed while maintaining comparable accuracy on the compound benchmark dataset \cite{zimmermann2019freihanddatasetmarkerlesscapture, zimmermann2017learningestimate3dhand}. While Guan \textit{et al.} \cite{lim2020mobilehand} remains the fastest, its model suffers a significant 42.8% drop in accuracy.
Our framework strikes a powerful balance between speed and accuracy. While SimpleHand\cite{zhou2024simplebaselineefficienthand} shows better accuracy, we achieve a substantial speed improvement, running at  15 FPS— 3.4 $\times$ faster. Likewise, we outperform MobileHand \cite{lim2020mobilehand} and MobRecon \cite{chen2022mobreconmobilefriendlyhandmesh} in accuracy (PA-MPJPE) and speed(FPS).

Unlike other transformer-based methods like FastMETRO \cite{lin2021end}, MeshGraphomer \cite{lin2021meshgraphormer}, and SimpleHand \cite{zhou2024simplebaselineefficienthand}, which struggle with real-time performance on edge devices, our model is highly efficient. While our model is not the most compact in terms of parameters, its architecture, which incorporates sparse convolution, is specifically optimized for efficiency.
For all models, inference time was measured on a Raspberry Pi 5 CPU and represents the mean over 50 repetitions. The number of parameters is listed in millions (M), and Frames Per Second (FPS) is calculated as the inverse of the forward pass time. This edge device-based evaluation clearly demonstrates the effective acceleration and real-world applicability of our model.\\

Table~\ref{tab:ablation_3d_decoding} presents a controlled unit test with identical input shapes. The proposed SPLite module achieves faster inference (averaged over 50 repetitions) while significantly reducing both FLOPs and parameter counts. This is accomplished by convolving only a quarter of the feature channels, thereby reducing memory access costs, combined with parallel vertex sampling to accelerate traversal sampling process in spiral convolution ~\cite{gong2019spiralnetfasthighlyefficient}. Together, these design choices deliver a substantial frame rate improvement over the baseline~\cite{chen2022mobreconmobilefriendlyhandmesh}.

Despite these optimizations, the overall parameter count of our end-to-end model remains substantial. This is because the primary operations of a typical encoding-decoding framework are in the 2D encoding stage. Since the ResNet-18 encoder is relatively large, the extensive reduction in our decoder's parameters has a limited impact on the total model size.

\begin{table}[h!]
    \centering
    \caption{The controlled unit test of our SPLite module w.r.t. SpiralConv++; Inference time(IF) is tested on Rasberry Pi 5 CPU, which is the forward time in the network;  \cite{zimmermann2019freihanddatasetmarkerlesscapture}.}
    \label{tab:ablation_3d_decoding}
    \begin{tabular}{l|cccc}
        \toprule
        3D decoding & Params(K) &FLOPs(M) $\downarrow$ &  IF(ms)$\downarrow$ \\
        \midrule
        SpiralConv~\cite{gong2019spiralnetfasthighlyefficient}++  & 21K          & 31.87     & 0.78 \\
        \midrule
        SPLite(ours)                                 & \textbf{5K}  & \textbf{9.36}   &\textbf{0.72}    \\
        \bottomrule
    \end{tabular}
\end{table}

\begin{table}[h!]
\centering
\caption{Accuracy on multimodality data after applying Quatization-Aware training }
\label{tab:QAT}
\begin{tabular}{l|cccc}
    \toprule
    precision (bits)& RGB PJ (mm)$\downarrow$& Edge PJ(mm)$\downarrow$ & Model size(MB)$\downarrow$  \\
    \midrule    
    FP32(w/o QAT)    & \textbf{9.0}& \textbf{11.2} & 71   \\
    INT8(w/ QAT)    & 9.1         & 12.3     & \textbf{18}         \\
    \bottomrule
\end{tabular}
\end{table}

\subsection{Qualitative Results}
In Figure~\ref{fig:wilddata1}, we evaluate our model's performance on challenging real-world images. The results highlight the limitations of the current fastest method by Lim \textit{et al.} \cite{lim2020mobilehand}, which struggles to produce accurate 3D hand pose estimations on such real-world data. In contrast, our approach significantly outperforms by predicting 3D hand poses that are much closer to the true hand configurations. The figure illustrates a side-by-side comparison: the input images, our model’s 3D joint estimations, 3D mesh reconstructions, and the outputs from MobileHand\cite{lim2020mobilehand} and Mobrecon\cite{chen2022mobreconmobilefriendlyhandmesh} for reference. Our method consistently generates more precise and anatomically plausible hand shapes and poses, demonstrating its robustness and superior generalization to wild, complex scenes.

In Figure~\ref{fig:wilddata2}, We processed real-world data using a proprietary algorithm to generate the edge images shown in the leftmost column. As illustrated in the figure, our method demonstrates superior accuracy in predicting the 3D hand joints and mesh. In contrast, existing approaches like MobileHand \cite{lim2020mobilehand} and Mobrecon \cite{chen2022mobreconmobilefriendlyhandmesh} appear to fail when presented with this different modality.

\begin{table}[h!]
\centering
\caption{Comparison of popular RGB-based real-world 3D hand datasets.}
\label{tab:dataset_comparison}
\begin{tabular}{l|ccccc}
\toprule
Dataset                                  & Size  & Mesh  & Multi-view  &Multi-modality \\
\midrule
 STB  ~\cite{zhang20163d}               & 36K   & $\times$ & $\times$ & $\times$  \\
 EgoDexter ~\cite{mueller2017real}      & 3K    & $\times$ & $\times$ & $\times$  \\
 Dexter+Object  ~\cite{sridhar2016real} & 3K     & $\times$ & $\times$ & $\times$ \\
 FreiHAND  ~\cite{zimmermann2019freihanddatasetmarkerlesscapture} & 134K  & $\checkmark$ & $\times$ & $\times$  \\
YoutubeHand ~\cite{kulon2020weakly}      & 47K & $\checkmark$ & $\times$ & $\times$  \\
HO3D  \cite{hampali2020honnotate}        & 77K & $\checkmark$ & $\times$ & $\times$ \\
DexYCB  \cite{chao2021dexycb}            & 528K & $\checkmark$ & $\times$ & $\times$  \\
H3D  \cite{yang2019aligning}              & 22K & $\checkmark$ & $\checkmark$(15) &  $\times$ \\
MHP \cite{gomez2019large}                 & 80K & $\checkmark$ & $\checkmark$(4) &  $\times$ \\
HOI4D  \cite{boukhayma20193dhandshapepose}  & 2.4M & $\checkmark$ &$\checkmark$  & $\times$  \\
HOT3D  \cite{boukhayma20193dhandshapepose}  & 1.16M & $\checkmark$ & $\checkmark$(4)  & $\times$  \\
\midrule
ours &135K  &$\checkmark$ & $\checkmark$(2) & $ \checkmark$(3)  \\
\bottomrule
\end{tabular}
\end{table}

In Table~\ref{tab:dataset_comparison}, we compare several prominent 3D hand pose datasets.
HOI4D~\cite{liu2024hoi4d4degocentricdataset} comprises 2.4 million egocentric video frames and is widely used in robotic training; it includes 16 object categories collected from 4 participants.  
HOT3D~\cite{banerjee2025hot3dhandobjecttracking}, captured using Meta's headset devices~\cite{engel2023projectarianewtool}, comprises 1.16 million well-annotated frames from 19 participants, spanning 33 object categories.  
In contrast, our dataset contains 135 thousand video frames collected from 20 participants, each interacting with 5 unique object categories—resulting in a total of 100 categories (Figure~\ref{fig:dataset}).  
Since object manipulation patterns vary across individuals, our dataset offers greater diversity in interaction styles, providing a valuable resource for robotics training.

%If your work needs an appendix, add it before the
%``\verb|\end{document}|'' command at the conclusion of your source
%document.

%tart the appendix with the ``\verb|appendix|'' command:
%\begin{verbatim}
%\end{verbatim}
%and note that in the appendix, sections are lettered, not
%numbered. This document has two appendices, demonstrating the %section
%and subsection identification method.

%The title, subtitle, keywords and abstract will be typeset in the main
%language of the paper.  The commands \verb|\translatedXXX|, \verb|XXX|
%begin title, subtitle and keywords, can be used to set these elements
%in the other languages.  The environment \verb|translatedabstract| is
%used to set the translation of the abstract.  These commands and
%environment have a mandatory first argument: the language of the
%second argument.  See \verb|sample-sigconf-i13n.tex| file for examples
%f their usage.

\section{conclusion}

In this work, we present a novel multimodal dataset and a lightweight deep learning framework designed to enhance hand-object interaction understanding, specifically targeting robotics and VR applications. Our dataset with around object categories and action manipulations outperforms existing ones  The proposed framework, utilizing a sparse convolution approach on a ResNet-18 backbone and the innovative SPLite decoder, achieves a significant performance boost, and send a significant signal for edge-device deployment. Through quantization-aware training, we maintain high accuracy while reducing memory usage. On a Raspberry Pi 5, our method delivers nearly a 3x speed-up compared to traditional methods, making it an efficient solution for real-time inference on edge devices.

Future research can explore further optimizations for edge device deployment, including the integration of more advanced compression techniques and hardware accelerators to enhance inference speed and energy efficiency. Additionally, expanding the dataset to include more diverse scenarios and interactions can further improve the robustness and generalization of models. We also plan to investigate the application of this framework in other domains, such as AR and teleoperation, where real-time performance and low-latency interactions are critical.

\section*{Acknowledgments}
This work is supported by the National Science and Technology Council (NSTC), R.O.C., under the projects “Advanced Technology of Intelligent Sensing Chip — Applied to Hand Pose Recognition System” and “NexIS: Next-Generation Intelligent Services through Self-Supervised and Trustworthy Learning Technologies.” We are also grateful for the support and guidance from Hsu Tzu Wei and Min Sun. Finally, we thank the anonymous reviewers for their valuable feedback.


{
    \small
    \bibliographystyle{unsrt}
    \bibliography{sample-base}
}

\appendix
% \begin{figure*}[t]
%     \begin{center}
%     \centering
%     \includegraphics[width=0.8\textwidth]{wild1.png}
%     \end{center}
%     \caption{Our method outperforms Mobrecon and MobileHand on several datasets in the wild}
%     \label{fig:wilddata1}

% \end{figure*}

% \begin{figure*}[t]
%     \centering
%     \includegraphics[width=0.8\textwidth]{wild2.png}
%     \caption{Our method is capable of performing inference on different data modalities, such as edge images, while others seem to fail}
%     \label{fig:wilddata2}
% \end{figure*}

\begin{figure*}[t]
    \begin{center}
    \centering
    \includegraphics[width=0.8\textwidth]{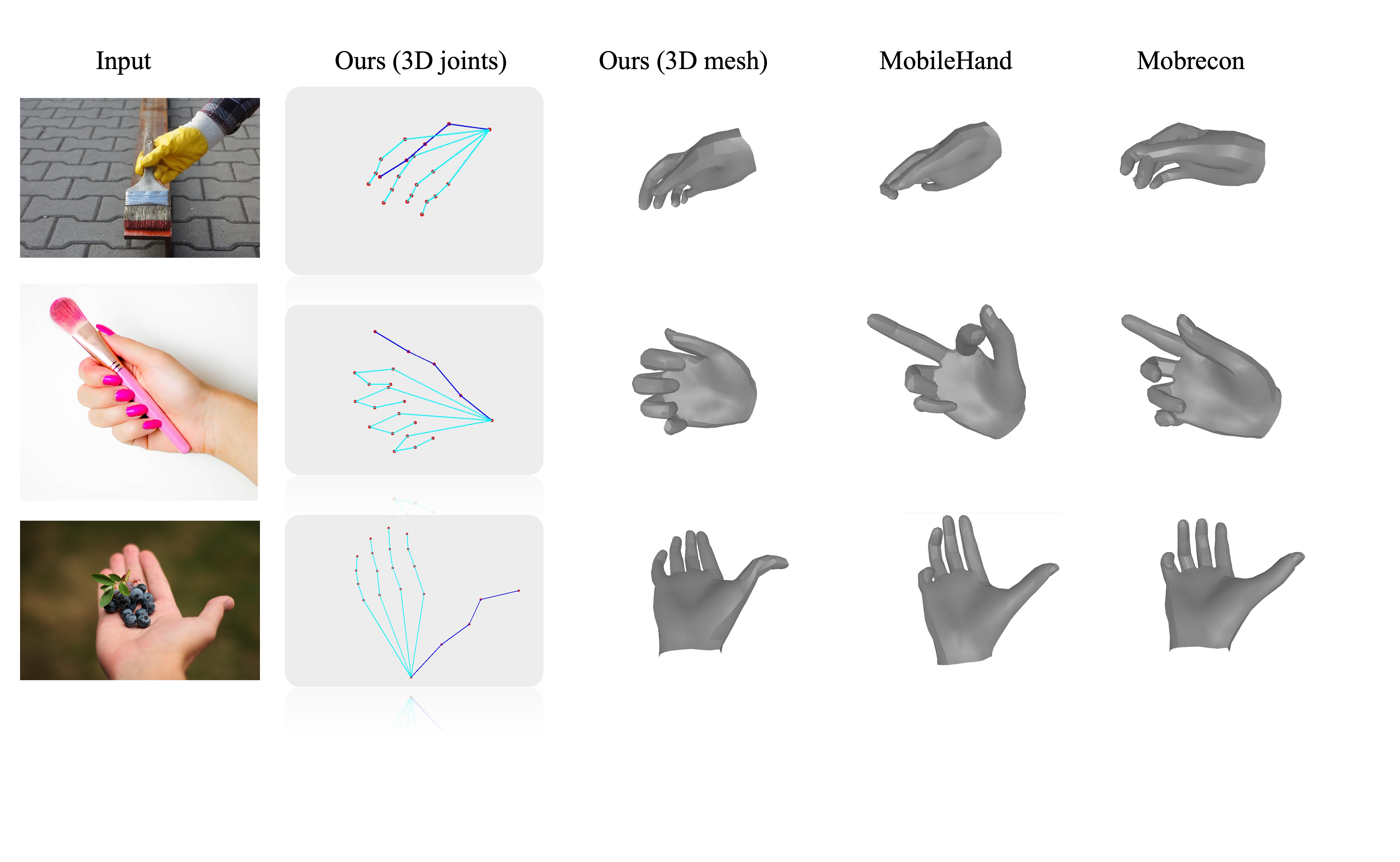}
    \end{center}
    \vspace{-50pt}
    \caption{Qualitative comparison on in-the-wild samples. Each row shows an input RGB image (left), followed by our predicted 3D joints, our reconstructed 3D mesh, and results from MobileHand\cite{lim2020mobilehand} and Mobrecon\cite{chen2022mobreconmobilefriendlyhandmesh}. Our method produces more accurate and natural hand poses, preserving fine articulation and finger alignment, while competing methods exhibit noticeable distortions or misalignments, particularly in challenging poses and occlusions.}
    \label{fig:wilddata1}

\end{figure*}
\begin{figure*}[t]
    \centering
    \includegraphics[width=0.8\textwidth]{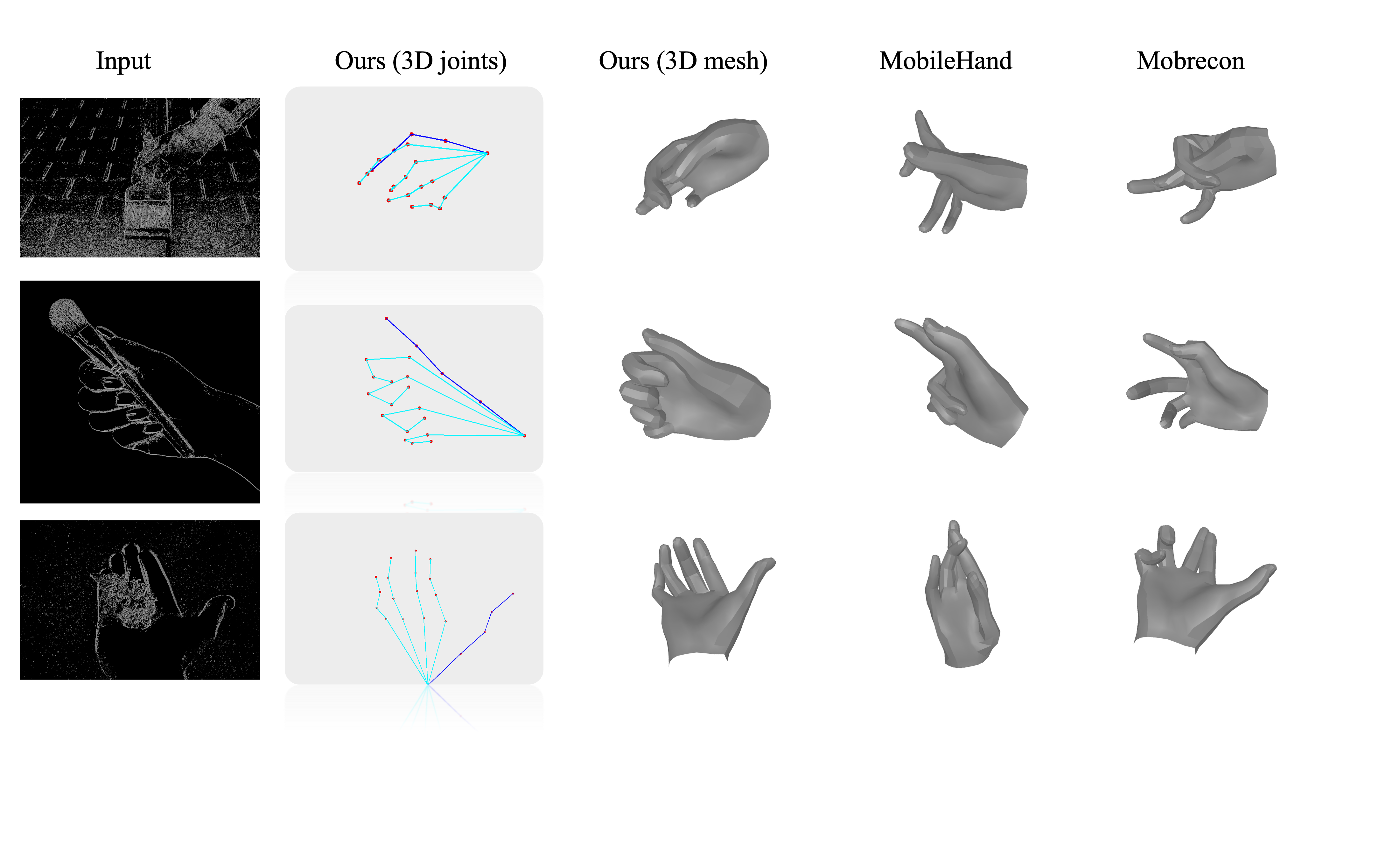}
    \vspace{-50pt}
    \caption{Qualitative comparison on in-the-wild samples. Each row shows an input edge image (left). Our method produces accurate and natural hand poses, while competing methods ~\cite{lim2020mobilehand} ~\cite{chen2022mobreconmobilefriendlyhandmesh} demonstrate unreasonable distortions and fail to produce a coherent hand structure on this modality.}
    \label{fig:wilddata2}
\end{figure*}
% \begin{figure}[t]
%   \centering
%   \includegraphics[width=0.9\linewidth]{edge_dataset.png}
%   \caption{The dataset is captured in real-world scenarios using a proprietary sensor, yielding 128x128 resolution images synchronized with two Intel RealSense cameras from different perspectives. The dataset includes four distinct scenarios, as shown in Figure~\ref{fig:dataset}. The ground-truth keypoints and meshes were labeled semi-manually.}
%   \label{fig:edgedataset}
% \end{figure}

\begin{figure*}[t]
    \centering
    \includegraphics[width=\linewidth]{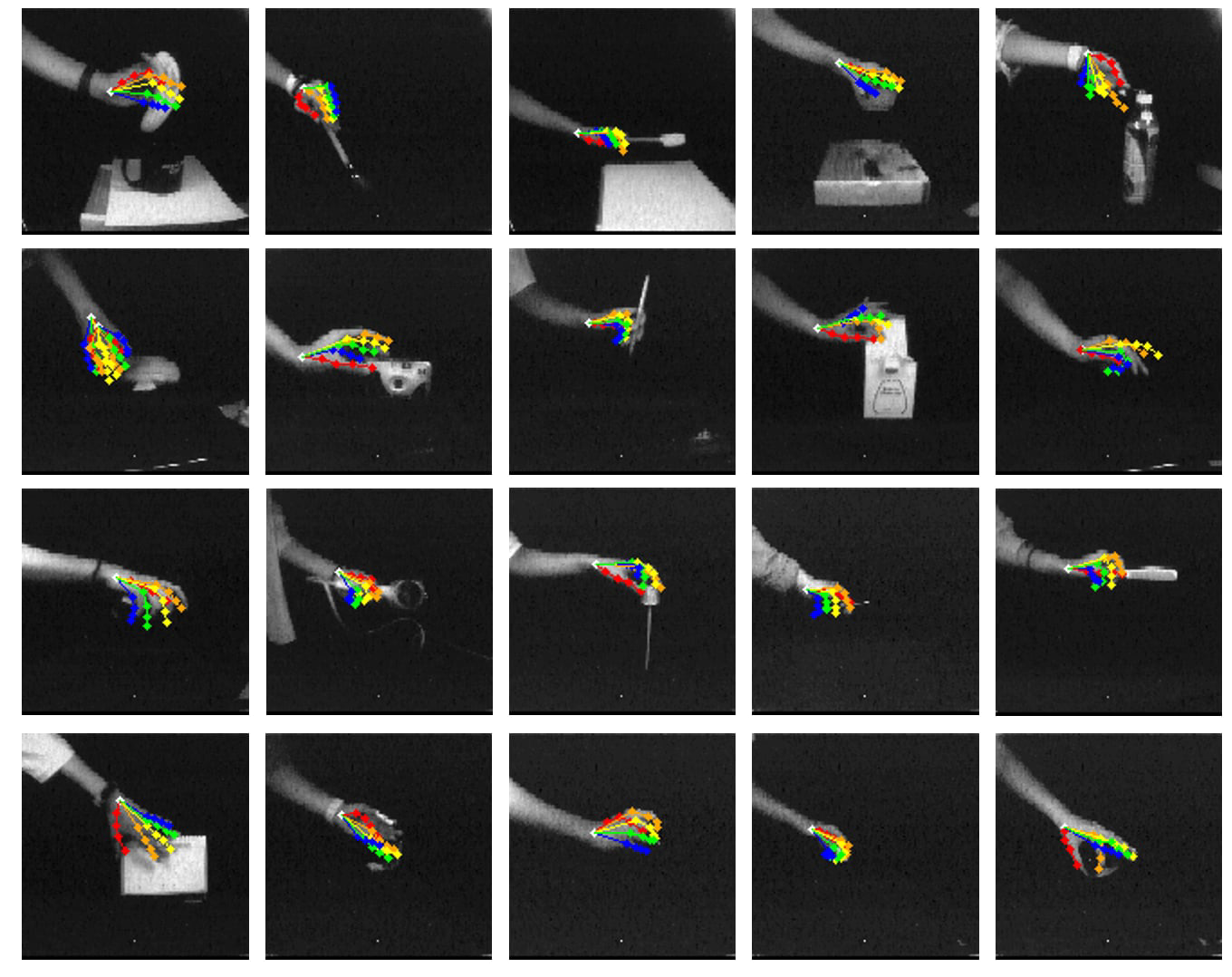}
    \caption{The dataset is captured in real-world scenarios using a proprietary sensor, yielding 128x128 resolution images synchronized with two Intel RealSense cameras from different perspectives. The dataset includes four distinct scenarios, as shown in Figure~\ref{fig:dataset}.100 genres of object categories and corresponding action manipulation respectively. The ground-truth keypoints and meshes were labeled semi-manually.}
    \label{fig:edgedataset}
\end{figure*}
\begin{figure*}[t]
  \centering
  \includegraphics[width=\linewidth]{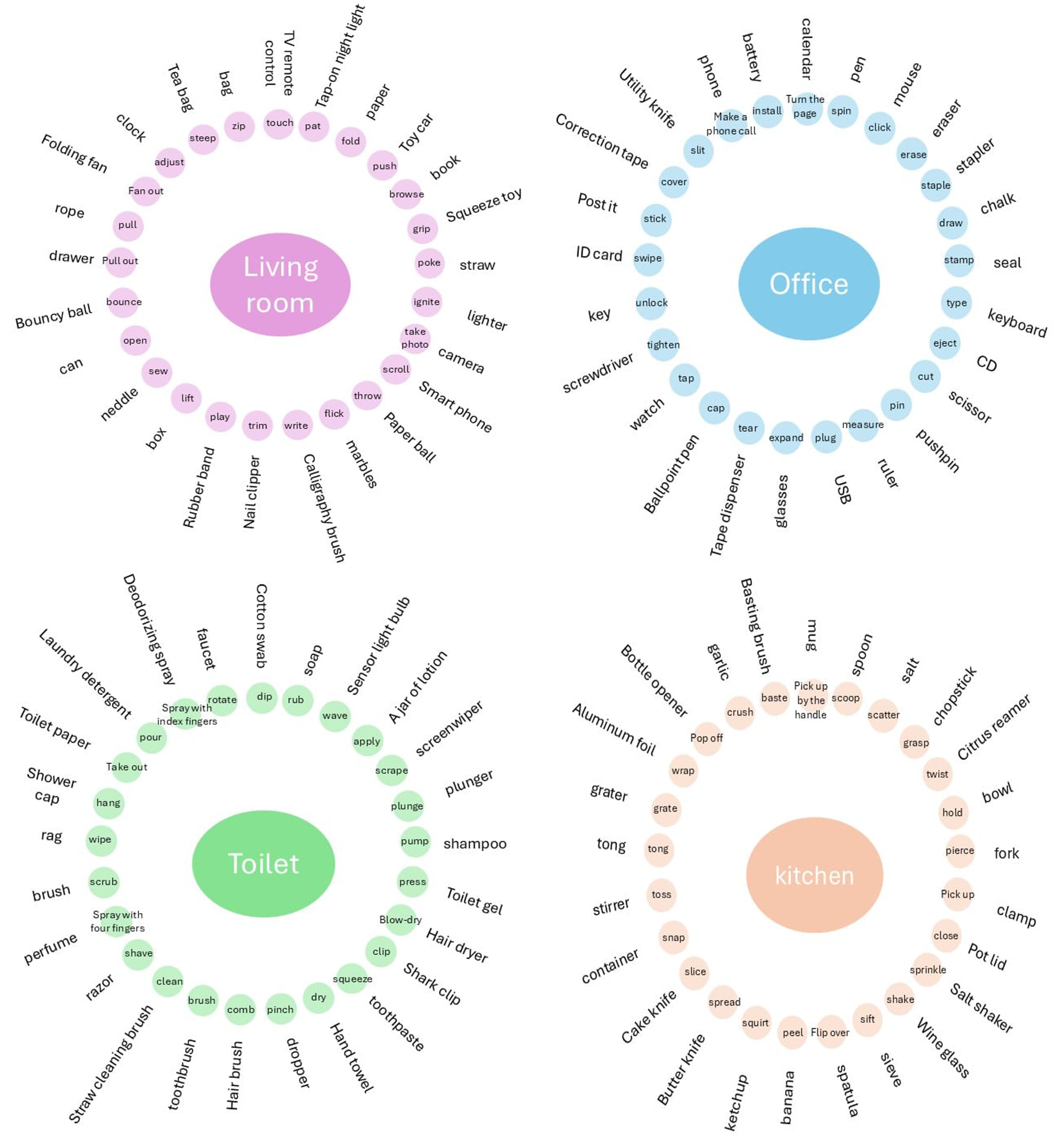}
  \caption{Illustration of our dataset, which includes object and motion categories—each with 25 unique objects and corresponding manipulation actions. We designed four common daily scenarios in the dataset to aid in robot model training. Images were captured using two Intel RealSense cameras and one proprietary camera. Ground-truth keypoints and meshes were semi-manually labeled, as detailed in the Appendix.}
  \label{fig:dataset}
\end{figure*}

\end{document}